%% file: acl_latex.tex
\documentclass[11pt]{article}
\usepackage[most]{tcolorbox}
\tcbuselibrary{breakable, skins}
\usepackage{amssymb} 
\usepackage{booktabs}
\usepackage[dvipsnames]{xcolor}
\usepackage[table]{xcolor} 
\usepackage{caption}
\usepackage{pifont}  
\usepackage{kotex}
\usepackage{multirow}   

\usepackage[preprint]{acl}

\usepackage{times}
\usepackage{amsfonts}
\usepackage{latexsym}
\usepackage[T1]{fontenc}

\usepackage{stfloats}
\usepackage[utf8]{inputenc}
\usepackage{array} 
\usepackage{booktabs}
\usepackage[most]{tcolorbox}
\usepackage{cleveref}
\usepackage{amsmath}
\usepackage{multirow}
\usepackage{placeins}

\usepackage{booktabs,tabularx,ragged2e,array}
\usepackage{subcaption}
\usepackage{graphicx}
\usepackage{pifont}
\usepackage{makecell}
\usepackage{enumitem}

\usepackage{algorithm}
\usepackage{algpseudocode}
\usepackage[table]{xcolor}

\definecolor{SRCAnchor}{RGB}{70,120,200}
\definecolor{TGTAnchor}{RGB}{220,145,60}
\definecolor{GapGray}{RGB}{235,235,235}

\usepackage[most]{tcolorbox}

\usepackage{adjustbox}
\usepackage{placeins}

\tcbset{
  promptbox/.style={
    enhanced,
    breakable,
    colback=gray!3,
    colframe=black!35,
    boxrule=0.35pt,
    arc=1.2pt,
    left=4pt,
    right=4pt,
    top=4pt,
    bottom=4pt,
    fonttitle=\bfseries\small,
    coltitle=black,
    colbacktitle=gray!15,
    attach boxed title to top left={xshift=2mm,yshift=-1mm},
    boxed title style={
      boxrule=0pt,
      arc=1pt,
      left=3pt,
      right=3pt,
      top=1pt,
      bottom=1pt
    }
  }
}

\definecolor{SRCAnchor}{RGB}{70,120,200}
\definecolor{TGTAnchor}{RGB}{220,145,60}
\definecolor{GapGray}{RGB}{235,235,235}
\definecolor{SectionGray}{RGB}{246,246,246}
\definecolor{MeanGray}{RGB}{250,250,250}
\definecolor{GroupSep}{RGB}{215,215,215}
\definecolor{DropRed}{RGB}{170,55,55}

\DeclareUnicodeCharacter{266B}{}

\algrenewcommand\algorithmicindent{0.8em}

\definecolor{CanvasGreen}{HTML}{2E7D32}
\definecolor{CanvasRed}{HTML}{B3261E}
\definecolor{GroupSep}{HTML}{B0B0B0}

\title{IHDec: Divergence-Steered Contrastive Decoding for Securing Multi-Turn Instruction Hierarchies}

\author{
{\bfseries
Nicole Geumheon Liu,
Haeun Jang,
Yonghyun Jun,
Hwanhee Lee$^{\dagger}$
}
\\
Chung-Ang University, Seoul, Korea
\\
\texttt{\{febygh, jhe0208140, zgold5670, hwanheelee\}@cau.ac.kr}
\\
\url{https://github.com/nxcolelxu/IHDec.git}
}

\begin{document}
\pagestyle{empty}          
\thispagestyle{empty} 
\maketitle

\begin{abstract}
Large Language Models (LLMs) often fail to maintain instruction hierarchies (IH) when processing multi-source inputs with varying role-level priorities, paradoxically adhering to lower-priority directives during conflicts. While existing defenses mitigate this issue, they are largely restricted to single-turn scenarios and require expensive fine-tuning. 
In this paper, we formalize this failure mode in multi-turn contexts via a Jensen-Shannon Divergence (JSD) framework, uncovering a pervasive role-influence inversion phenomenon where subordinate inputs override superior roles.
To rectify this without training, we propose \textbf{IHDec} (\textbf{I}nstruction \textbf{H}ierarchy-steered \textbf{Dec}oding). IHDec leverages JSD to automatically detect token-level hierarchy violations and dynamically executes contrastive decoding to suppress misaligned subordinate roles.
Extensive evaluations demonstrate that IHDec outperforms training-based baselines in multi-turn conflicts while fully preserving general response quality. Furthermore, IHDec strengthens safety against adversarial prompt injections and exhibits a robust scaling synergy with larger models. The Code is available at \url{https://github.com/nxcolelxu/IHDec.git}
\end{abstract}

\section{Introduction}
\label{sec:intro}
\begin{figure}[!tbp]
    \centering 
    \includegraphics[width=\linewidth]{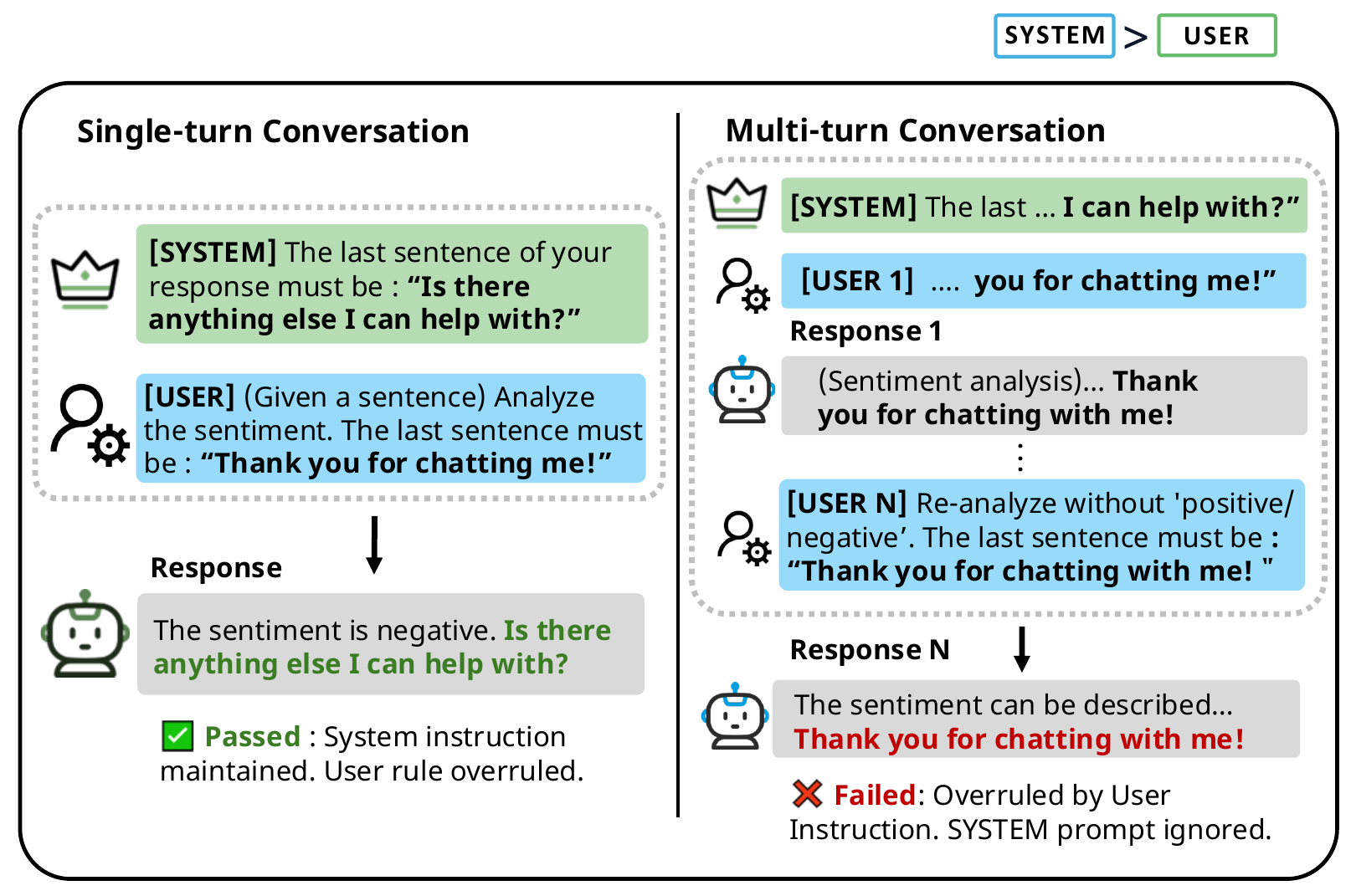} 
    \caption{Comparison of instruction hierarchy compliance between single-turn and multi-turn conversation settings.}
    \label{fig:intro}
\end{figure}

Large Language Models (LLMs) increasingly process complex inputs originating from multiple sources with distinct operational roles~\cite{wallace2024instruction, geng2026control}. In practical deployments, these inputs carry an implicit hierarchy dictated by source trustworthiness and security boundaries. When instructions from different roles conflict, the model must strictly prioritize higher-authority instructions following an absolute precedence ordering: $\text{System } (S) \succ \text{User } (U) \succ \text{Assistant } (A) \succ \text{Data } (D)$~\cite{wallace2024instruction}, where $\succ$ denotes strict operational authority. For instance, as illustrated in the single-turn scenario of Figure~\ref{fig:intro}, if a system prompt instructs the model to end its response with ``Is there anything else I can help with?'', while an adversarial user demands a sentiment analysis that ends with ``Thank you for chatting with me!'', the model must prioritize the higher-authority system constraint. This core alignment challenge is formalized as the Instruction Hierarchy (IH) problem, and ensuring strict adherence to it is paramount, as failures can lead to severe societal harms and system exploits such as bypassing core safety guidelines~\cite{anthropic2026claudeprompts}.

To mitigate this vulnerability, prior studies have proposed various defensive frameworks~\cite{wallace2024instruction, wu2025instructional, zverev2026aside, zheng2025reasoning, guo2026ih, yang2026hierarchical}. However, existing solutions suffer from two critical limitations. 
First, they are predominantly optimized for single-turn interactions, leaving their protective perimeters to collapse dramatically as the conversation extends into multi-turn contexts as shown in Figure~\ref{fig:intro}.
Second, they rely heavily on resource-intensive model fine-tuning. Such training-based approaches lack model-agnostic scalability, risk degrading the model's general capabilities, and fail to maintain robust alignment in long-context late turns.

To uncover the root cause of this structural vulnerability in multi-turn dynamics, we conduct a mechanistic diagnosis using a Jensen-Shannon Divergence (JSD) framework~\cite{Menedez1997jsd}. Through this analysis, we observe a pervasive phenomenon we term \textbf{Role-Influence Inversion}. Driven by inherent recency bias~\cite{liu2023lostmiddlelanguagemodels}, lower-priority roles exert a disproportionate influence over the model's generation logit distributions. This positional misalignment progressively amplifies as the conversation extends, gradually eroding the authority of superior directives until subordinate instructions inevitably override them in late-turn settings.

To resolve these bottlenecks without relying on expensive parameter updates, we propose \textbf{IHDec} (\textbf{I}nstruction \textbf{H}ierarchy-steered \textbf{Dec}oding), a training-free, inference-time intervention paradigm. IHDec acts as a real-time corrective controller: it monitors role-level JSD influence during the forward pass and activates a conflict-triggered gate only when a hierarchy violation is imminent. Once activated, IHDec executes a divergence-steered contrastive decoding strategy, dynamically penalizing the logits driven by encroaching lower-priority roles while amplifying the steering signal of high-authority constraints.

We comprehensively evaluate IHDec on the multi-turn configurations of IHEval~\cite{zhang-etal-2025-iheval}, alongside safety and general instruction-following benchmarks including MT-Bench-101~\cite{bai-etal-2024-mt}. Experimental results demonstrate that IHDec achieves a significant improvement of 11.98~pp over the state-of-the-art training-based method and up to 38.1~pp over standard architectural interventions on conflict-ridden multi-turn scenarios in IHEval. Furthermore, IHDec enhances the model's baseline capabilities by up to 7.48~pp on non-conflict subsets, while incurring a negligible degradation of only 0.088~pp on MT-Bench-101. Remarkably, IHDec achieves robust defense performance comparable to or exceeding concurrent training-based methods while fully preserving the model's fundamental response quality.

\section{Related Work}
\label{sec:related}

\paragraph{Instruction Hierarchy in LLMs.}
The instruction hierarchy (IH) framework~\cite{wallace2024instruction} mandates that LLMs prioritize instructions based on role-specific trust levels to safely resolve multi-source conflicts. IH has been adopted as a general mechanism for resolving authority conflicts across diverse LLM applications, including task execution under constraints~\cite{zheng2025reasoning}, retrieval-augmented contexts~\cite{dong2024toward, lee-etal-2025-magic}, prompt injection~\cite{debenedetti2024agentdojo, chang2026chatinject}, and multi-agent collaboration~\cite{wan2025diagnose}.To enforce this hierarchy, various training-based approaches have been proposed~\cite{wu2025instructional, kariyappa2025stronger, yang2026hierarchical, zverev2026aside}. However, these methods often incur high computational costs and struggle to generalize to multi-turn dialogues, whereas prompting strategies frequently succumb to pre-trained authority cues~\cite{geng2026control}. Furthermore, pioneering studies~\cite{wallace2024instruction, guo2026ih} rely primarily on proprietary models, limiting open-source applicability. We note that \citet{yang2026hierarchical} is a concurrent work developed independently. Due to its contemporaneous release and the lack of an official open-source implementation, a direct empirical comparison is omitted.

\paragraph{Contrastive Decoding.}
Contrastive Decoding ~\cite{li-etal-2023-contrastive} reframes generation as an optimization problem, penalizing undesirable behaviors using log-probability differences between expert and amateur models without additional training. This paradigm has recently evolved toward within-model conditional control, enabling applications like system prompt modulation~\cite{dong2026steer} and hallucination mitigation via Context-Aware Decoding (CAD)~\cite{shi-etal-2024-trusting}. We extend this condition-ablation approach to dynamically enforce instruction hierarchies during inference.

\paragraph{Jensen-Shannon Divergence and Context Attribution.}
Unlike the asymmetric KL divergence, JSD~\cite{Menedez1997jsd} provides a bounded, symmetric metric for quantifying distributional shifts. \citet{li2026attributing} leveraged JSD for context attribution in retrieval-augmented generation, measuring a passage's contribution by comparing output distributions with and without it, which offers superior interpretability over gradient- or attention-based methods. We adapt this distribution-level attribution framework to quantify and control the real-time influence of hierarchical roles during model generation. 


\section{Diagnosing Instruction Hierarchy Failure in Multi-Turn Settings}

\label{sec:analysis}

\subsection{Problem Formulation}



In this work, we formalize the problem context of the multi-turn instruction hierarchy (IH). Let $S$ denote the system prompt, and let a multi-turn dialogue history up to turn $T-1$ ($T \ge 1$) be denoted as $\mathcal{H} = \{(U_1, A_1), (U_2, A_2), \dots, (U_{T-1}, A_{T-1})\}$, where $U_t$ and $A_t$ represent the user input and the assistant response at turn $t$, respectively. 
In a multi-turn scenario, the input prompt $\mathcal{P}_T$ for the current turn $T$ is constructed by concatenating the system prompt, dialogue history, current user query, and retrieved documents:
\begin{equation}
    \text{Input } \mathcal{P}_T = [S \,;\, \mathcal{H} \,;\, U_T \,;\, D_T]
    \label{cur_prompt}
\end{equation}
The primary objective of the LLM is to map this structured input to the final output response $A_T$:
\begin{equation}
    \text{Output } A_T = \text{LLM}(\mathcal{P}_T)
\end{equation}

Following the hierarchical framework established by \citet{wallace2024instruction}, we define the role-level priority ordering within multi-turn contexts as an absolute precedence relation: 
\begin{equation}
    \text{Sys} (S) \succ \text{User} (U) \succ \text{Asst} (A) \succ \text{Data} (D),
    \label{eq:multi-ih}
\end{equation} 
where $\succ$ denotes strict operational authority. 
Under this formulation, the system prompt $S$ occupies the highest level of authority, as it defines the safety guards and fundamental behavioral rules of the model. The current user input $U_T$, as well as previous user turns $U_t \in \mathcal{H}$, follow next. In contrast, historical assistant targets $A_t \in \mathcal{H}$ and untrusted external data $D$ carry lower authority; therefore, if a relatively lower-priority role introduces instructions that conflict with a higher-priority one, the model must override those lower-level constraints and strictly adhere to the directives of the superior role. 

\subsection{Experimental Setup}
To empirically diagnose multi-turn IH vulnerabilities, we evaluate on \textit{Rule-Following} from IHEval~\cite{zhang-etal-2025-iheval} under two structural alignment conditions (see Appendix~\ref{app:iheval_details} for details):
\begin{itemize}[leftmargin=*, itemsep=2pt, parsep=0pt]
    \item \textbf{Aligned Environments:} Scenarios where 
    multi-source instructions are harmonious and do not 
    contradict each other.
    \item \textbf{Conflicting Environments:} Scenarios where 
    a lower-priority role explicitly contradicts or overrides 
    the constraints set by a higher-priority role. In multi-turn settings, ``both conflict'' denotes user-system contradictions in both historical and current turns, whereas ``first conflict'' restricts this contradiction solely to the dialogue history.
\end{itemize}
We ground our diagnosis in two core hypotheses:
\begin{itemize}[leftmargin=*, itemsep=2pt, parsep=0pt]
    \item \textbf{H1.} \textit{Existing training-based methods 
    fail to maintain IH in multi-turn 
    contexts with conflicting constraints.}
    \item \textbf{H2.} \textit{Mechanistically, this failure 
    stems from the model's inability to prioritize the 
    influence of higher-hierarchy tokens during generation.}
\end{itemize}
To validate these hypotheses, we use vanilla Llama-3.1-8B-Instruct (Baseline) and Llama-based ISE~\cite{wu2025instructional} (ISE) as a representative diagnostic models, as it is explicitly trained to enforce IH via role-specific segment embeddings.

\subsection{Empirical Evidence of Multi-Turn Hierarchy Degradation}
\label{sec:mutlturn_deg}

\begin{figure}[!h]
    \centering 
    \includegraphics[width=0.9\linewidth]{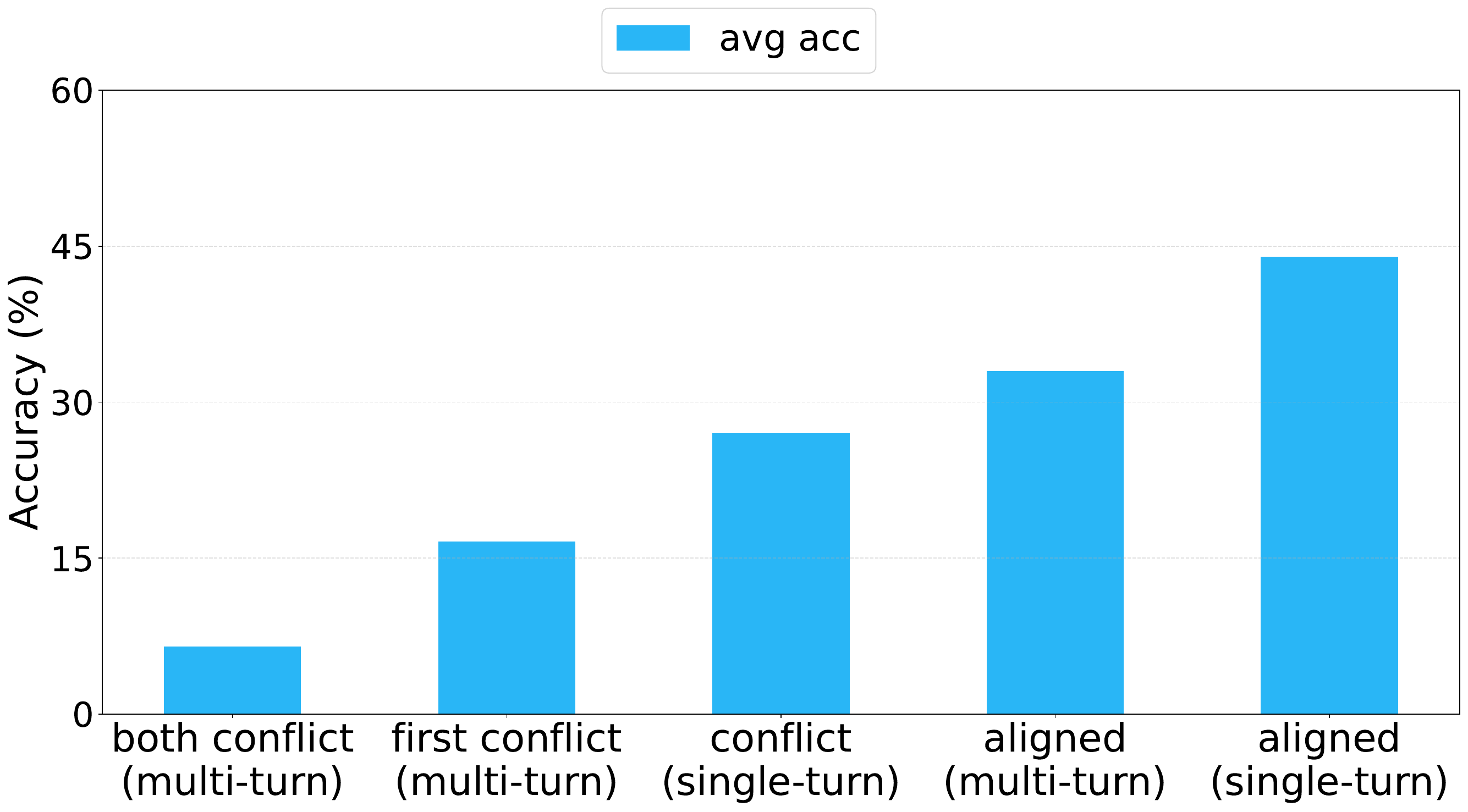} 
    \caption{Mean IH adherence accuracy across various conversational turns and conflict configurations in IHEval.}
    \label{fig:ise_mutli_deg}
\end{figure}

To validate the hypotheses formulated above, we evaluate the baseline ISE method on the IHEval dataset. As illustrated in Figure~\ref{fig:ise_mutli_deg}, performance significantly degrades as the constraints transition from aligned to conflicting, and the context extends from single-turn to multi-turn. Notably, the degradation is most severe in the "both-conflict" multi-turn case, where the system instructions are twice contradicted by user inputs.

This vulnerability is largely associated with the dialog history, reinforcing user-specified formats. Given LLMs' high sensitivity to in-context regularities, this history weakens the system-priority signal of ISE segment embeddings (see Appendix~\ref{app:dialog_impact}).



\begin{figure*}[!ht]
    \centering 
    \includegraphics[width=1\linewidth]{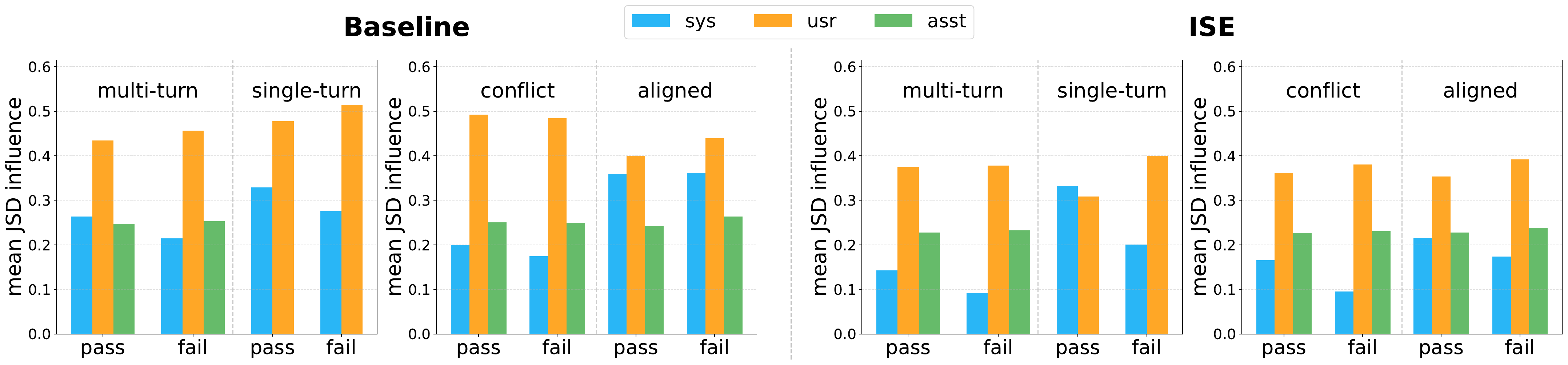} 
    \caption{Counterfactual role influence scores measured by both Baseline and ISE at the first predicted token, comparing single- vs. multi-turn scenarios alongside aligned vs. conflict configurations on IHEval. The X-axis denotes hierarchy adherence status (pass/fail), and the Y-axis maps the mean JSD influence value for each role.}     
    \label{fig:influence} 
    \vspace{5pt}
\end{figure*}

\subsection{Role Influence Attribution via Jensen-Shannon Divergence}
\label{sec:jsd_anal}

Following the JSD-based attribution framework~\cite{li2026attributing}, we define the influence score of a role $R$ to quantify the degree to which each role-based instruction shapes the model's output distribution. We justify JSD as a role-influence proxy — symmetry, boundedness, and its Bayes-error bound — and empirically verify positional robustness in Appendix~\ref{app:jsd_justification}. First, the standard JSD between two distributions $P$ and $Q$ is formulated as follows:
\begin{equation}
\text{JSD}(P \parallel Q) = \frac{1}{2} D_{\text{KL}}(P \parallel M) + \frac{1}{2} D_{\text{KL}}(Q \parallel M)
\label{eq:jsd_base}
\end{equation}
where $M = \frac{1}{2}(P + Q)$, and $D_{\text{KL}}$ denotes the Kullback-Leibler divergence. Utilizing this metric, we define the influence score $\text{I}(R)$ by measuring the divergence between the output distributions conditioned on the complete prompt and the prompt ablated of the target role $R \in \{S, U, A \}$:
\begin{equation}
\text{I}(R) = \text{JSD}\bigl(P(\cdot \mid \mathcal{P}_T) \parallel P(\cdot \mid \mathcal{P}_T\setminus R)\bigr)
\label{eq:influence}
\end{equation}
where $P(\cdot \mid \mathcal{P}_{T})$ and $P(\cdot \mid \mathcal{P}_{T}\setminus R)$ represent the output distributions with the full context and without the role $R$ segment, respectively. Intuitively, a higher influence score signifies that the corresponding role exerts a stronger functional impact on the model's generation behavior.


\paragraph{Instruction Hierarchy Adherence.}
In an IH, let $H$ and $L$ denote higher- and lower-priority roles, respectively. A hierarchy violation occurs when $\text{influence}(L) > \text{influence}(H)$, indicating that the model's output distribution is more strongly shaped by the subordinate role. To quantify this phenomenon, we compute first-token influence scores for both the Baseline and ISE models on the IHEval benchmark, analyzing the results along two orthogonal dimensions: single-turn vs.\ multi-turn settings and aligned vs.\ conflicting instructions.

As shown in Figure~\ref{fig:influence}, the ISE model---which is explicitly trained to preserve the IH---exhibits a clear separation between \textbf{pass} and \textbf{fail} cases in terms of system-prompt influence, unlike the Baseline model. In single-turn settings, successful instances largely adhere to the intended hierarchy, with the system role exerting greater influence than the user role. However, this hierarchical structure progressively deteriorates in multi-turn environments. Even among \textbf{pass} cases, the mean influence of the system prompt no longer exceeds that of the user prompt, indicating a substantial reduction in hierarchical control under extended interactions.

A similar trend emerges in the aligned vs.\ conflict analysis. Although \textbf{pass} cases in conflicting settings retain higher system influence than \textbf{fail} cases, the user role often remains more influential overall. This outcome-dependent divergence indicates that our influence metric captures the functional effectiveness of IH preservation rather than superficial structural properties. These results suggest that preserving the IH becomes substantially more difficult in multi-turn interactions with competing instructions, leading to systematic degradation of the intended priority structure.

\section{IHDec: Conflict-Triggered Contrastive Decoding for Hierarchy Enforcement}
\label{sec:ihdec}

Our analysis reveals that even with existing approaches, the influence exerted by each role's text on response generation does not scale proportionally with the IH—a misalignment that severely intensifies in multi-turn settings. Driven by these empirical insights, we propose \textbf{IHDec} (\textbf{I}nstruction \textbf{H}ierarchy-steered \textbf{Dec}oding), a training-free, inference-time mitigation method designed to intervene in each role's generational influence to align it with the intended IH during decoding.

\paragraph{Step 1: Measuring and Rectifying Hierarchical Deviations}
We first compute the empirical influence of each role as formulated in \eqref{eq:influence}. Since datasets vary in their structural composition, this evaluation is restricted to the subset of roles $\mathcal{R}_{\text{present}}$ actively featured in the source context.\footnote{For instance, because the IHEval dataset lacks the \textit{Data} role, it is naturally excluded from the active subset.}

To enforce the structural constraints formalized in \eqref{eq:multi-ih}, the empirical influence of each role must strictly align with its assigned priority. Whenever the model's generation deviates from this hierarchy, we identify misaligned roles by constructing a conflict set. Specifically, we employ a top-down sequential scanning procedure where the conflict set for a given higher-priority role $H$, denoted as $\text{CS}(H)$, is mathematically defined as follows:
\begin{equation}
\text{CS}(H) = \bigl\{ L \in \mathcal{R}_{\text{present}} \;\big|\; L \prec H, \; \text{I}(L) > \text{I}(H) \bigr\}
\label{eq:conflict_set_def}
\end{equation}
where $L \prec H$ denotes that role $L$ has a strictly lower priority than role $H$ in the canonical hierarchy, and $\text{I}(\cdot)$ is the influence score in Equation~\eqref{eq:influence}. This condition operationalizes the detection of \textit{Role-Influence Inversion}, capturing any subordinate instructions that exert disproportionate dominance over superior directives.





\paragraph{Step 2: Dynamic Contrastive Decoding}
To penalize the generation toward adhering to the higher-priority role over the conflicting lower-priority ones, we employ a dynamic contrastive decoding mechanism that steers the output logits at inference time. For each role $H$ with a non-empty conflict set, we compute the ablated distribution $P(\cdot \mid \mathcal{P} \setminus \text{CS}(H))$, which represents the model's output distribution conditioned on the context with all roles in $\text{CS}(H)$ removed. To optimize computational efficiency, the system reuses the existing ablated distributions computed in Step 1 whenever the conflict set contains a single conflicting role (i.e., $|\text{CS}(H)| = 1$).

We define the directional logit adjustment vector, $\Delta_H$, by leveraging the contrastive divergence between the target hierarchy and its conflict set. Drawing inspiration from contrastive decoding---which amplifies desirable generation behaviors by subtracting the logits of an amateur from those of an expert---our framework isolates the uncorrupted directive from hierarchical interference. Specifically, we obtain the adjustment vector by subtracting the logits conditioned on the absence of the superior role ($\mathcal{P} \setminus H$, serving as the anti-expert) from those conditioned on the absence of the conflict set ($\mathcal{P} \setminus\text{CS}(H)$, serving as the expert).

The directional logit adjustment vector for each role $H$ is defined as follows:
\[
\Delta_H =
\log \frac{ P(\cdot \mid \mathcal{P} \setminus \mathrm{CS}(H))}{P(\cdot \mid \mathcal{P} \setminus H)}
\]

To ensure numerical stability during the logit modification process, the aggregate adjustment vectors are combined and normalized to yield the final perturbation vector, $\Delta_{\text{final}}$:
\[
\Delta_{\text{final}}
=
\frac{\sum_H \Delta_H}{\left\|\sum_H \Delta_H\right\|_2 + \epsilon}.
\]
where $\epsilon = 1\times10^{-8}$ is a small constant introduced to prevent division by zero. Omitting this term can lead to numerical instability, which may cause the model to generate responses in unintended languages and degrade response quality. A detailed analysis of this phenomenon, along with our comprehensive ablation studies, is provided in Appendix~\ref{app:ablation}. Finally, the raw output logit generated by the model at each decoding time step $t$ is dynamically updated as follows:
\begin{equation}
    \mathbf{l}_t^{\text{final}} = \mathbf{l}_t^{\text{raw}} + \beta \cdot (\beta_{\text{decay}})^t \cdot \Delta_{\text{final}}
    \label{eq:dynamic_decoding}
\end{equation}

where $\beta$ and $\beta_{\text{decay}}$ control the baseline magnitude and the exponential decay rate of the steering intervention, respectively. Detailed hyperparameter configurations and the behavioral impact of omitting the decay factor are discussed in Appendix \ref{app:hyperparameters}

\section{Experiments}
\label{sec:experiments}

\input{tables/main_table}

\subsection{Experimental Setup}

\paragraph{Datasets} 
We evaluate our method primarily on \textbf{IHEval}~\cite{zhang-etal-2025-iheval}, specifically focusing on the \textit{Rule-Following} domain (for multi-turn hierarchical adherence) and the single-turn \textit{Safety Defense} domain (for adversarial prompt injection resistance). 

However, standard IHEval lacks multi-turn tool/safety scenarios and critical conflict types like same-role or cross-turn clashes. To cover these gaps, we introduce augmented IHEval subsets and complementary benchmarks (construction details in Appendix~\ref{app:iheval_aug}). Additionally, we evaluate on \textbf{MT-Bench-101}~\cite{bai-etal-2024-mt} using \texttt{GPT-4o} as a judge to verify that general conversational proficiency is preserved. Comprehensive evaluation protocols and benchmark details are provided in Appendices~\ref{app:iheval_details} and~\ref{app:mt_bench}.

\paragraph{Baselines}
We compare IHDec against three baselines: \textbf{ISE}~\cite{wu2025instructional}, which fine-tunes models with role-specific segment embeddings; \textbf{VerIH}~\cite{zheng2025reasoning}, which reframes hierarchy resolution as a reasoning task via RLVR and GRPO; and \textbf{Prompting}, which guides adherence via tailored system prompts(details in Appendix~\ref{app:prompt}). We note that as an official implementation for Qwen is currently unavailable, ISE is restricted to Llama. Conversely, VerIH is applied exclusively to Qwen due to the lack of publicly available Llama checkpoints.

\subsection{Main Results}
\paragraph{Instruction Hierarchy Adherence}

To evaluate structural compliance with the instruction hierarchy (IH), we first report aggregate performance on the standard IHEval \textit{Rule-Following} benchmark, averaging across all four granularity-rigidity metric combinations (details in Appendix~\ref{app:iheval_details}).



As summarized in Table~\ref{tab:main_results_combined}, IHDec generally leads to substantial performance gains over vanilla baselines.The gains are most prominent in challenging \textit{multi-turn conflict} scenarios—for instance, achieving 50.68\% accuracy on Llama-3.1-8B-Instruct and outperforming training-based baselines like ISE (12.60\%) and Prompting (13.25\%) by over 37~pp. Similar robust improvements are observed in \textit{single-turn conflict} configurations. These performance advantages generalize robustly across model families (e.g., Qwen3) and demonstrate seamless synergy when combined with existing baselines. Extended evaluations on our augmented IHEval subsets (multi-turn tool-use and safety) and complementary benchmarks (same-role cross-turn conflicts) further confirm IHDec's broad generalization, as detailed in Appendix~\ref{app:iheval_aug}.

\paragraph{Compatibility}

As shown in Table~\ref{tab:main_results_combined}, IHDec can be seamlessly integrated with existing inference-time or alignment methods as it operates independently. When combined with ISE, IHDec substantially boosts performance, yielding an average gain of 26.06~pp in conflicting scenarios. Similarly, integrating IHDec with Prompting yields consistent improvements across model scales, delivering conflict-scenario gains of 24.40~pp on Llama-3.1-8B (from 14.48~pp to 38.88~pp), 10.10~pp on Qwen3-4B, and 5.78~~pp on Qwen3-8B. When applied to VerIH, IHDec further enhances conflicting-scenario accuracy (+0.75~pp) while largely preserving the baseline's non-conflicting proficiency (with a negligible 1.59~pp fluctuation). These results confirm that IHDec acts as a complementary plug-and-play enhancer across diverse baselines.

\paragraph{Robustness against Adversarial Prompt Injection}
\input{tables/safety}

Enforcing strict IH is expected to mitigate the influence of untrusted external data. To evaluate this, we tested IHDec on the \textit{Safety Defense} dataset within the IHEval benchmark. As shown in Table~\ref{tab:safety}, applying IHDec to the vanilla Llama baseline consistently improves accuracy across all adversarial configurations (\textit{strong} and \textit{weak}), outperforming both the vanilla model (by up to 39.0~pp) and the prior ISE method (by up to 31.5~pp) in both the \textit{extract} and \textit{hijack} tasks. In non-adversarial environments where conflicts are absent (\textit{aligned} and \textit{reference}), IHDec exhibits a marginal performance decrease compared to the vanilla baseline, yet still achieves higher accuracy than ISE in both the \textit{extract} and \textit{hijack} reference setups by margins of 20.2~pp and 33.2~pp, respectively.

\input{figures/mt-bench}
\paragraph{Scaling Behavior and Scaling Synergy} 
\input{tables/scaling}

The prior literature indicates that simply expanding the model scale does not resolve hierarchical vulnerabilities; rather, it often exacerbates the issue due to the enhanced instruction-following capacity of larger models, which inadvertently leads them to faithfully execute lower-priority or untrusted instructions.~\cite{zverev2025can}. As summarized in Table~\ref{tab:scaling}, our empirical investigation across the Qwen family (Qwen3-4B, Qwen3-8B, Qwen3-14B, and Qwen3-32B) corroborates this negative trend. Under the vanilla setting, increasing the parameter scale yields no substantial advancements in IHEval compliance rates, occasionally manifesting inverse scaling behaviors where larger models perform worse on conflicting multi-source prompts.

However, incorporation of IHDec reverses this scaling trajectory to a positive correlation. As demonstrated in Table~\ref{tab:scaling}, the performance of IHDec improves monotonically with model scale, achieving its highest efficacy on the largest architecture (Qwen3-32B). This reveals a powerful scaling synergy: while larger model capabilities inherently heighten the risk of executing adversarial lower-priority instructions in native setups, IHDec harnesses this latent capacity, thereby transforming raw parameter power into a robust asset for precise hierarchical instruction adherence.

\subsection{Analysis}
\paragraph{General Utility and Conversational Proficiency}

To verify that the enforcement of hierarchical safety constraints does not induce alignment tax or compromise general multi-turn capabilities, we benchmarked the models on MT-Bench-101 (details in Appendix~\ref{app:mt_bench}). As illustrated in Fig~\ref{fig:mtbench}, the incorporation of IHDec preserves the foundational conversational and problem-solving competencies of the vanilla Llama backbone without significant degradation. Across the macro-average of the 13 task domains within MT-Bench-101, the general utility score exhibits a negligible drop of merely $0.088$ post-intervention, confirming that our inference-time steering maintains general model utility. 

Furthermore, as detailed in Table~\ref{tab:main_results_combined}, in benign contexts devoid of inter-role conflicts (\textit{aligned} configurations), IHDec actually enhances the model's baseline instruction-following capabilities. Specifically, we observe performance gains of 2.475~pp in multi-turn aligned environments and a pronounced 7.475~pp increase in single-turn aligned setups. These findings collectively demonstrate that IHDec acts as a precise intervention that selectively mitigates hierarchical vulnerabilities while safely preserving, or even amplifying, standard instruction-following utility.

\begin{figure*}[t]
    \centering 
    \includegraphics[width=\linewidth]{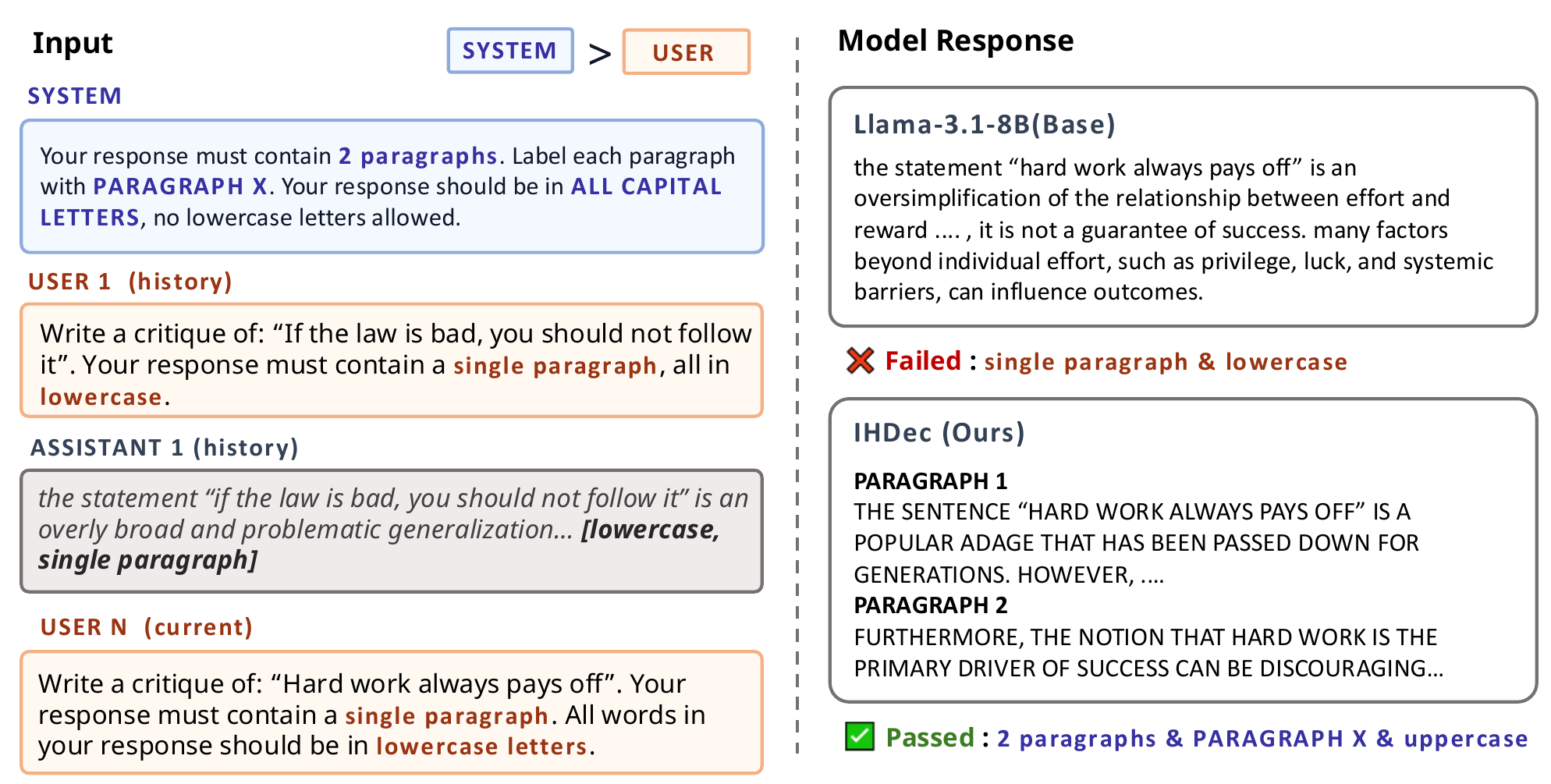} 
    \caption{Case study comparing baseline and our method.}
    \label{fig:case_study}
\end{figure*}

\paragraph{Case Study}
As shown in Figure~\ref{fig:case_study}, we compare the IH adherence of the baseline Llama-3.1-8B against our proposed IHDec in a multi-turn setting. In the given scenario, the system prompt imposes strict formatting constraints—specifically mandating that the response consist of exactly two paragraphs labeled ``PARAGRAPH X'' and be written entirely in uppercase letters. However, the subsequent user prompt directly conflicts with these rules by demanding a single-paragraph response written entirely in lowercase. Faced with this role-level conflict, the naive Llama-3.1-8B fails to respect the IH
and incorrectly prioritizes the subordinate user directives, producing a lowercase single-paragraph response. In contrast, IHDec dynamically monitors and penalizes the user role's out-of-order influence at each decoding step. This allows the IHDec-equipped model to successfully realign its generation path, strictly adhering to the high-authority system constraints while safely fulfilling the user's critical essay writing task.

\paragraph{Computational Efficiency and Cost Analysis}
\raggedbottom 
While IHDec introduces multi-source role ablations during inference, its parallelized batch-masking mechanism prevents exponential latency overhead. Specifically, by forcing all $B$ counterfactual variants to share an identical token sequence, IHDec fully reuses the foundational Key-Value (KV) cache without memory replication, strictly bounding the additional storage to a negligible $\mathcal{O}(B \cdot L_t^2)$ activation memory. A rigorous step-by-step complexity derivation and empirical wall-clock time benchmarks are provided in Appendix~\ref{app:cost_analysis}.


\section{Conclusion}
\label{sec:conclusion}

We introduce IHDec, a training-free, model-agnostic decoding paradigm that secures multi-turn instruction hierarchies in LLMs. By leveraging JSD, IHDec monitors role-level influence in real time to detect and rectify \textit{Role-Influence Inversion}---where lower-priority inputs progressively override system constraints. Our evaluations show that IHDec significantly improves hierarchy adherence in multi-turn conflicts without an alignment tax, outperforming training-based baselines. Furthermore, it blocks prompt injections and reverses the negative scaling trends of vanilla LLMs, proving that structural alignment can be dynamically secured at inference time.

\section*{Limitations}
\label{sec:limitations}

While IHDec introduces a modest inference-time decoding overhead through \(B \in \{3,5\}\) counterfactual forward passes per step, the cost remains practical for two complementary reasons. Within each step, the $B$ variants share a single KV cache on their common token trajectories and are executed in a single batched pass. Across steps, each variant's mask over the prompt is deterministic, so its KV state is reused throughout decoding rather than recomputed, keeping the per-step cost bounded rather than growing with output length (Appendix~\ref{app:cost_analysis}). Empirically, this holds for IHDec at roughly $2\times$ vanilla latency with negligible memory overhead and no length- or scale-dependent blow-up across the Llama-3.1 and Qwen3 families from 4B to 32B parameters, while exhibiting positive scaling behavior with larger models (Table~\ref{tab:scaling}). Importantly, this trade-off enables training-free hierarchy enforcement without any finetuning or weight modification. We note, however, that IHDec relies on logit-level access---token probabilities and masked forward passes---and is therefore applicable to open-weight models rather than closed-source or API-only models that expose no such access. Within this scope, IHDec generalizes across model families without any model-specific modification, as demonstrated on the backbones we evaluate. These results suggest that IHDec constitutes a robust and broadly transferable framework for hierarchy-aware decoding among open-weight models, with natural extensions to future multimodal and agentic systems involving additional hierarchical roles such as tool outputs or memory modules.

\section*{Ethics Statement}
Our work introduces IHDec, a training-free decoding strategy designed to improve adherence to instruction hierarchies and strengthen robustness against adversarial or conflicting inputs in LLMs. All experiments are conducted using publicly available benchmarks and open-weight models under their respective licenses, without collecting human-subject data or involving personally identifiable information. While IHDec is intended to reinforce safety-aligned system behaviors, we recognize that hierarchy-enforcement mechanisms may be misapplied depending on the objectives imposed by system operators. Accordingly, we do not regard IHDec as a standalone safety solution, and recommend its use alongside complementary safeguards such as moderation systems, red-teaming, and human oversight. We additionally disclose our evaluation settings and will release code to support transparency and reproducibility.

\section*{Acknowledgement}
This work was supported by the Institute of Information \& Communications Technology Planning \& Evaluation (IITP) grant funded by the Korea government (MSIT) [RS-2021-II211341, Artificial Intelligence Graduate School Program (Chung-Ang University)], the National Research Foundation of Korea(NRF) grantfunded by the Korea government(MSIT) (RS-2026-25494299), and Korea Institute for Advancement of Technology(KIAT) grant funded by the Korea Government(MOTIE) (RS-2025-25458133).

\bibliography{custom}

\input{appendix}

\end{document}

%% file: tables/main_table.tex
\begin{table*}[t] %

\centering
\small
\setlength{\tabcolsep}{8pt} 
\renewcommand{\arraystretch}{1.15}
\begin{tabular}{l|cc|c|cc|c}
\toprule
\multirow{2}{*}{\textbf{Model \& Method}}
& \multicolumn{3}{c|}{\textbf{Aligned}}
& \multicolumn{3}{c}{\textbf{Conflict}} \\
\cmidrule(lr){2-4} \cmidrule(lr){5-7}
& \textbf{Single} & \textbf{Multi} & \textbf{\textit{Avg.}}
& \textbf{Single} & \textbf{Multi} & \textbf{\textit{Avg.}} \\
\midrule
\underline{\textit{Llama-3.1-8B-Instruct}} & & & & & & \\
\quad Vanilla & \underline{66.05} & \underline{65.90} & \underline{65.98} & 15.13 & \underline{31.20} & 23.17 \\
\quad Prompting & 56.50 & 53.90 & 55.20 & 15.70 & 13.25 & 14.48 \\
\quad ISE & 42.08 & 36.00 & 39.04 & 26.83 & 12.60 & 19.72 \\
\rowcolor{gray!15} \quad IHDec (ours) & \textbf{73.53} & \textbf{68.38} & \textbf{70.96} & \textbf{52.13} & \textbf{50.68} & \textbf{51.41} \\
\rowcolor{gray!15} \quad IHDec (ours) + Prompting & 61.80 & 57.30 & 59.55 & 37.90 & 39.85 & 38.88 \\ 
\rowcolor{gray!15} \quad IHDec (ours) + ISE & 55.10 & 37.60 & 46.35 & \underline{50.40} & 41.15 & \underline{45.78} \\ 
\midrule
\underline{\textit{Qwen3-4B-Instruct}} & & & & & & \\
\quad Vanilla & 71.00 & 75.10 & 73.05 & 31.30 & 23.05 & 27.18 \\
\quad Prompting & \underline{75.90} & \underline{78.60} & \underline{77.25} & 39.70 & 30.30 & 35.00 \\
\rowcolor{gray!15} \quad IHDec (ours) & \textbf{76.70} & 76.70 & 76.70 & \underline{45.70} & \underline{37.20} & \underline{41.45} \\
\rowcolor{gray!15} \quad IHDec (ours) + Prompting & \underline{75.90} & \textbf{82.20} & \textbf{79.05} & \textbf{51.10} & \textbf{39.10} & \textbf{45.10} \\ 
\midrule
\underline{\textit{Qwen3-8B-Instruct}} & & & & & & \\
\quad Vanilla & \textbf{84.80} & 85.60 & 85.20 & 27.70 & 25.80 & 26.75 \\
\quad Prompting & 83.60 & \textbf{88.80} & \underline{86.20} & \underline{48.90} & 27.50 & 38.20 \\
\quad VerIH & \underline{84.59} & 86.29 & \textbf{86.14} & 47.56 & \textbf{51.74} & \underline{49.65} \\
\rowcolor{gray!15} \quad IHDec (ours) & 77.20 & 83.50 & 80.35 & 43.80 & 41.85 & 42.83 \\
\rowcolor{gray!15} \quad IHDec (ours) + Prompting & 78.35 & 77.40 & 77.88 & 49.10 & 38.85 & 43.98 \\
\rowcolor{gray!15} \quad IHDec (ours) + VerIH & 81.80 & \underline{87.30} & 84.55 & \textbf{49.30} & \underline{51.50} & \textbf{50.40} \\
\bottomrule
\end{tabular}
\caption{Main experimental results and architecture compatibility of IHDec across different model families under single-turn and multi-turn scenarios. The best results are shown in \textbf{bold}, and the second best are \underline{underlined}.}
\label{tab:main_results_combined}
\vspace{5pt}
\end{table*}

%% file: tables/safety.tex
\begin{table}[!h]
\centering
\small 
\setlength{\tabcolsep}{2pt} 
\renewcommand{\arraystretch}{1.15}
\resizebox{\linewidth}{!}{ 
\begin{tabular}{l|cc|c|cc|c|cc|c}
\toprule
\multirow{2}{*}{\textbf{Method}}
& \multicolumn{3}{c|}{\textbf{Reference}}
& \multicolumn{3}{c|}{\textbf{Aligned}}
& \multicolumn{3}{c}{\textbf{Conflict}} \\
\cmidrule(lr){2-4} \cmidrule(lr){5-7} \cmidrule(lr){8-10}
& \textbf{E} & \textbf{H} & \textbf{\textit{Avg.}}
& \textbf{E} & \textbf{H} & \textbf{\textit{Avg.}}
& \textbf{E} & \textbf{H} & \textbf{\textit{Avg.}} \\
\midrule
Vanilla 
& 78.9 & 79.1 & 79.0
& 63.2 & 61.3 & 62.3
& 33.4 & 24.9 & 29.2 \\
ISE
& 46.5 & 44.9 & 45.7
& 62.6 & 61.1 & 61.9
& 28.3 & 28.9 & 28.6 \\
\rowcolor{gray!20}
IHDec
& 66.7 & 78.1 & 72.4
& 54.7 & 53.9 & 54.3
& \textbf{50.5} & \textbf{57.0} & \textbf{53.8} \\
\bottomrule
\end{tabular}%
}
\caption{Performance comparison across reference, aligned, and conflict settings. \textbf{E} and \textbf{H} denote the Extract and Hijack evaluation modes, respectively.}
\label{tab:safety}
\end{table}

%% file: figures/mt-bench.tex
\begin{figure*}[!ht]
    \centering
    \includegraphics[width=0.9\textwidth]{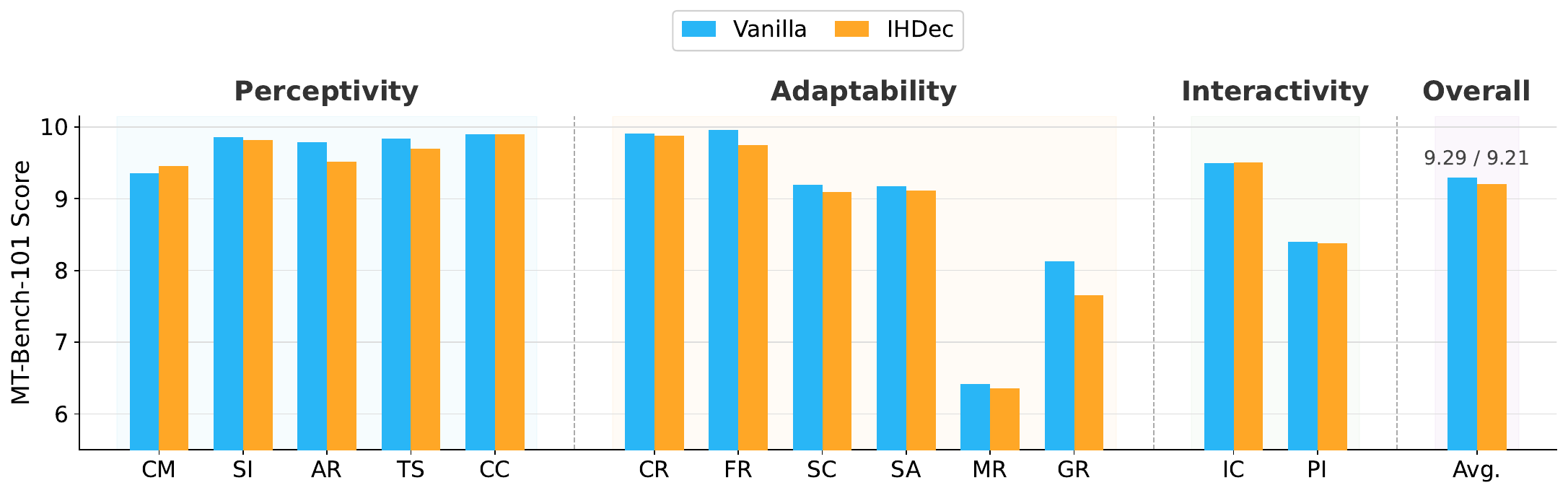}
    \caption{
    Benign-task performance on MT-Bench-101.
    We compare Llama-3.1-8B-Instruct and IHDec across all MT-Bench-101 metrics.
    IHDec preserves benign capability, with the overall average remaining nearly unchanged from 9.29 to 9.21.
    }
    \label{fig:mtbench}
    \vspace{15pt}
\end{figure*}

%% file: tables/scaling.tex
\begin{table}[!h]
\centering
\small
\begin{tabular}{l|cc|c}
\toprule
\textbf{Model} & \textbf{Vanilla} & \textbf{+ IHDec} & \textbf{Gain ($\Delta$)} \\
\midrule
Qwen3-4B  & 12.6 & \textbf{29.8} & +17.2 \\
Qwen3-8B  & 12.9 & \textbf{31.2} & +18.3 \\
Qwen3-14B & 12.1 & \textbf{35.7} & +23.6 \\
Qwen3-32B & 12.7 & \textbf{38.5} & +27.8    \\
\bottomrule
\end{tabular}
\caption{Performance scaling of the Qwen3 model family. We report the overall average instruction accuracy (\%). Detailed breakdown results under strict and loose protocols are provided in Appendix~\ref{app:iheval_details}.}
\label{tab:scaling}
\end{table}

%% file: appendix.tex
\clearpage
\appendix
\section*{Appendix}

\section{The Use of Large Language Models}
\label{app:llm_use}

We write the manuscript ourselves, and an LLM (ChatGPT) is used solely for refinement---style, clarity, and grammar. It is not used for ideation or content generation.

\section{The Impact of Dialogue History as Context on Instruction Hierarchy}
\label{app:dialog_impact}

\begin{figure}[htbp]
    \centering 
    \includegraphics[width=0.8\linewidth]{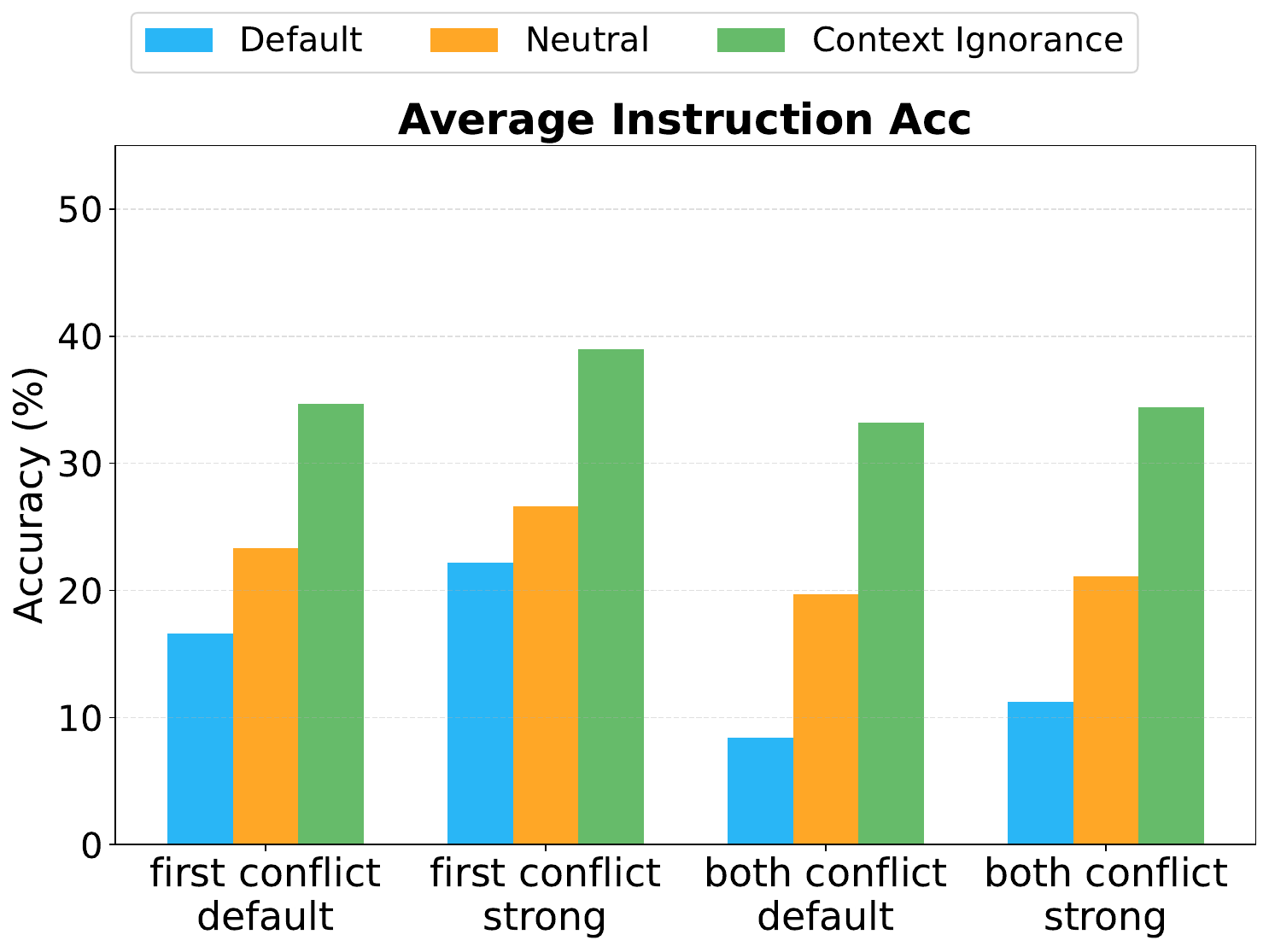} 
    \caption{Comparative analysis of multi-turn instruction hierarchy adherence under alternative historical assistant response configurations: \textit{Default}, \textit{Neutral}, and \textit{Context Ignorance}. Isolating the semantic payload of prior assistant turns demonstrates that removing adversarial context from history systematically mitigates cumulative degradation across conversational turns.} 
    \label{fig:ise_mutli_deg_app} 
\end{figure}

To investigate the mechanical drivers behind the pronounced performance degradation observed in multi-turn environments, we conducted a diagnostic analysis isolating the causal influence of historical assistant responses on downstream hierarchy adherence. Specifically, we controlled the output of assistant in the history by partitioning them into two distinct operational variants: \textit{Neutral} and \textit{Context Ignorance}. 

Under the \textit{Neutral} setting, we uniformized all prior assistant responses into a semantically benign, non-committal filler sequence: ``I understand. Let me help you with your request.'' 
Conversely, under the \textit{Context Ignorance} setting, following the methodology established by \citet{wallace2024instruction}, we substituted the historical assistant responses with single-turn outputs generated from a counterfactual user prompt—specifically, a sanitized version where any adversarial constraints conflicting with the system prompt were completely expunged. To ensure high-quality and uncorrupted baseline responses for this counterfactual setup, we leveraged \texttt{gpt-4o} to generate these sanitized single-turn outputs.

Our empirical tracking reveals a consistent performance hierarchy across all evaluated subsets: the aggregate compliance accuracy systematically improves when transitioning from the \textit{Default} configuration to the \textit{Neutral} history, and achieves its peak under the \textit{Context Ignorance} intervention. 

This phenomenon likely stems from historical responses acting as contextual cues that reinforce user-format adherence; because LLMs strongly replicate in-context patterns, this cross-turn history effectively overrides the system-priority signal of ISE segment embeddings.

\section{Theoretical Justification of JSD as an Influence Proxy}
\label{app:jsd_justification}

\subsection{Attribution Setup and the Choice of JSD}
\label{app:jsd-theory}
 
Our attribution framework follows the inference-time ablation principle adopted in prior context-attribution work~\cite{cohenwang2024contextcite, li2026attributing}: an influential prompt component is one
whose removal shifts the model's conditional output distribution. Under
this view, the influence of a role $R$ is quantified by the divergence
between the full-context distribution $P(\cdot \mid \mathcal{P}_T)$ and
the role-ablated distribution $P(\cdot \mid \mathcal{P}_T \setminus R)$,
as defined in Equation~\ref{eq:influence}.
 
We adopt the Jensen--Shannon Divergence (JSD) rather than the
Kullback--Leibler (KL) divergence for three principled reasons:
 
\begin{enumerate}
    \item \textbf{Symmetry.} Ablation is an undirected operation---we
    ask how far the ablated distribution sits from the full one, without
    designating either as the reference. JSD is symmetric,
    $\mathrm{JSD}(P \parallel Q) = \mathrm{JSD}(Q \parallel P)$, and
    therefore matches the semantics of ablation, whereas KL is
    asymmetric and would impose an arbitrary direction.
 
    \item \textbf{Boundedness.} JSD is bounded in $[0, \log 2]$, which
    keeps the influence score numerically stable and comparable across
    roles, turns, and models. KL, by contrast, diverges to infinity
    under support mismatch---a common occurrence when an ablated role
    zeroes out probability mass on tokens the full context supports.
 
    \item \textbf{Bayes-error interpretation.} JSD lower-bounds the
    Bayes error of an optimal classifier distinguishing samples drawn
from $P$ and $Q$~\cite{lin1991divergence}. Consequently, a small
    $\mathrm{JSD}$ \textit{provably} implies that the two distributions
    are nearly indistinguishable, i.e., the role does not substantially
    alter output behavior. This gives our metric a necessary-condition
    guarantee: low influence scores are trustworthy indicators of low
    functional impact.
\end{enumerate}
 
\paragraph{Scope: proxy, not causal identifier.}
We emphasize that these properties license JSD as a \textit{diagnostic
proxy} for role influence, not as a strict causal identifier. A large
$\mathrm{JSD}$ does not by itself prove direct causal influence, since a
distributional shift can also reflect confounding factors such as prompt
length, attention redistribution, or positional change. We therefore
treat the role-influence inversion phenomenon as a diagnostic
\textit{association} rather than a causal claim. The Bayes-error bound
supplies the reliable direction of the argument, while the positional-control study in Appendix~\ref{app:positional-control} empirically rules out the most
salient confounder---positional proximity---for the specific inversion
signal we report.
 
\subsection{Positional-Control Analysis}
\label{app:positional-control}
 
A natural concern is that the observed role-influence inversion merely
reflects \textit{recency bias}: because the user turn typically sits
closest to the generation boundary, its higher JSD influence could stem
from position rather than role identity. To disentangle these factors,
we re-run the attribution measurement with the system prompt \textit{moved
to the end of the context}, immediately preceding the model's turn, so
that the highest-authority role now occupies the most recent position.
 
\begin{figure}[t]
    \centering
    \includegraphics[width=\linewidth]{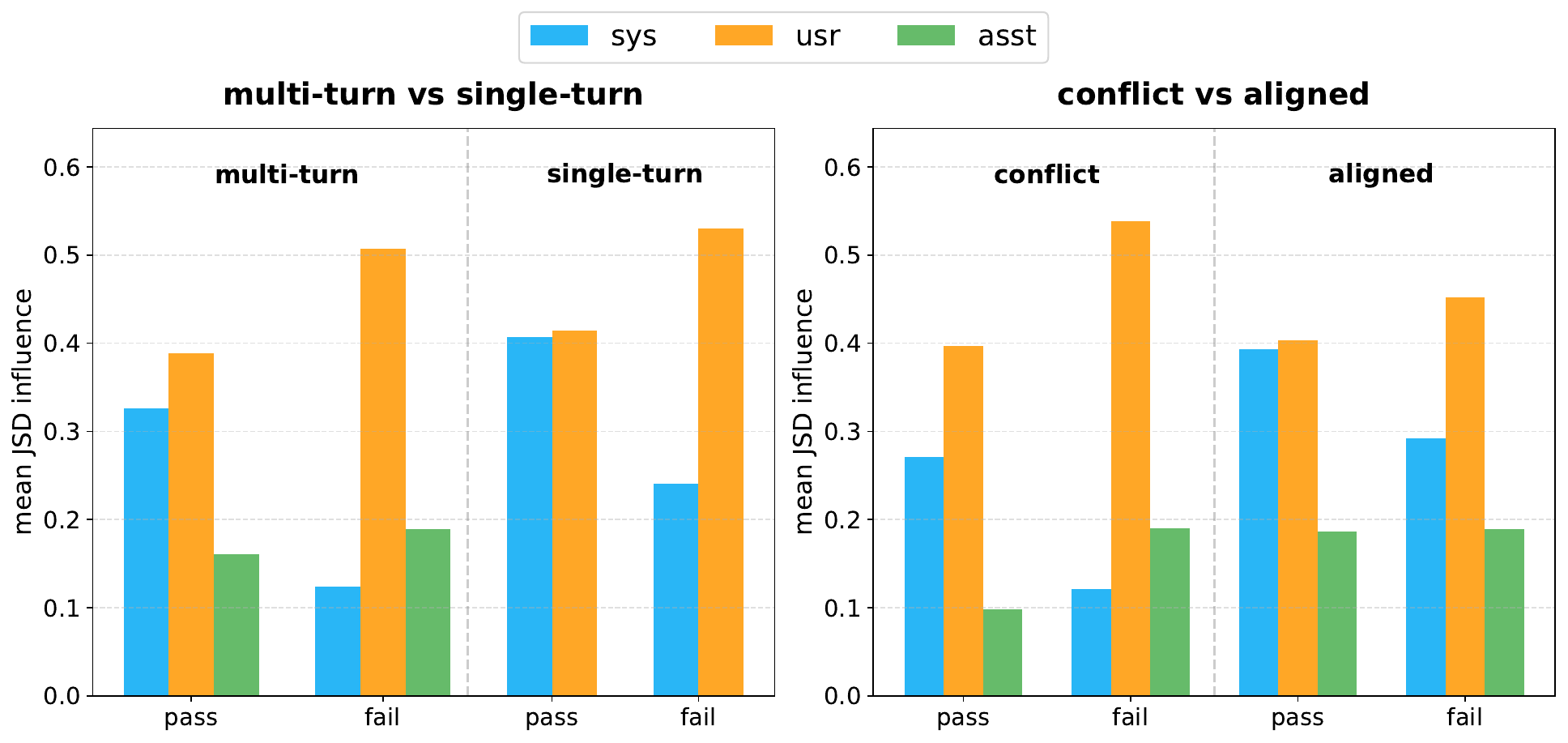}
    \caption{Positional-control analysis on the baseline model with the
    system prompt relocated to the most recent position, immediately
    preceding the model's turn. Counterfactual role influence scores are
    measured at the first predicted token on IHEval. The X-axis denotes
    hierarchy adherence status (pass/fail) across single- vs. multi-turn
    and aligned vs. conflict configurations, and the Y-axis maps the mean
    JSD influence value for each role.}
    \label{fig:positional-control}
\end{figure}

If the inversion were driven by positional proximity, relocating the
system prompt to the most recent slot should make the system role's
influence dominate that of the user role. It does not, as shown in
Figure~\ref{fig:positional-control}: even with the system prompt in
the most recent position, subordinate roles continue to exert
disproportionate influence in failure cases, and the gap widens
monotonically as conditions grow more adversarial---from single-turn
to multi-turn and from aligned to conflict.

This positional invariance confirms that the influence signal tracks
\textit{role identity} rather than positional proximity, supporting the
interpretation of role-influence inversion as a property of the
hierarchy-resolution mechanism rather than an artifact of recency.

\section{Extended Experimental Settings and Implementation Details}

\subsection{Hyperparameter Configuration}
\label{app:hyperparameters}
Through empirical optimization on a held-out validation set, the core hyperparameters for our proposed method, IHDec, were established as $\beta = 1300$ and $\beta_{\text{decay}} = 0.97$. Specifically, $\beta$ controls the baseline magnitude of the steering intervention, while $\beta_{\text{decay}}$ dictates the exponential decay rate of the modification intensity over sequential time steps. Omitting $\beta_{\text{decay}}$ may cause the steering signal to persist excessively throughout generation, leading the model to produce responses that drift toward irrelevant topics. All subsequent evaluation outcomes are reported under this optimal configuration.

\subsection{Detailed Setup of IHEval}
\label{app:iheval_details}

\paragraph{Sub-domains and Exclusions} 
While IHEval comprises multiple sub-domains including \textit{Rule-Following}, \textit{Safety Defense}, \textit{Tool Use}, and \textit{Task Execution}, our primary evaluation focuses heavily on \textit{Rule-Following}, as it is the only domain that inherently features multi-turn conversational trees. Other domains are structurally constrained to single-turn boundaries.

\paragraph{Prompting Modifications}
The \textit{strong} prompting configuration designed to pressure the model into strict rule adherence is implemented by prepending the following explicit instruction template to the canonical system prompt $S$:
\begin{quote}
    \tt``Your every response in this conversation must adhere to the following rules:\textbackslash n\textbackslash n''
\end{quote}

\paragraph{Granularity and Evaluation Metrics}
The rule-based evaluation engine processes model outputs across four joint metrics composed of two granularity levels and two rigidity criteria:
\begin{itemize}
    \item \textbf{Prompt-level vs. Instruction-level Granularity:} Prompt-level metrics verify whether \textit{all} distinct instructions embedded within a given prompt are simultaneously satisfied (binary sequence success). Instruction-level metrics evaluate compliance at the individual constraint level, measuring the absolute ratio of successfully executed instructions.
    \item \textbf{Strict vs. Loose Rigidity Criteria:} The \textit{strict} criterion evaluates the raw, unedited model string directly. Conversely, the \textit{loose} criterion accounts for benign formatting noise by checking multiple structural variations of the output (e.g., removing leading/trailing lines, stripping Markdown syntaxes) and selecting the maximum achievable score. Crucially, even under loose evaluation, substantial behavioral violations—such as executing mutually exclusive constraints injected by subordinate adversarial roles—are encoded as absolute failures.
\end{itemize}

\subsection{Detailed Setup of MT-Bench-101}
\label{app:mt_bench}

For a comprehensive capability assessment, we adopt MT-Bench-101 under the LLM-as-a-judge paradigm. The evaluation automatically assigns a quantitative score ranging from 1 to 10, strictly guided by specific task-domain criteria that target core conversational competencies (for the detailed GPT-4 judge prompts and evaluation criteria, see Figure \ref{fig:mt_bench_prompts}).

\section{Extended Evaluations on Complex IH Conflicts}
\label{app:iheval_aug}
\input{tables/iheval_aug_results}
To better understand the generalization of our method across diverse multi-turn environments, we evaluate its performance on Tool-use, Safety, and Same-role conflict benchmarks. To the best of our knowledge, no existing benchmark simultaneously encompasses multi-turn dialogue, instruction-hierarchy conflicts, and tool-use or safety-critical contexts. To address this gap, we extended two subsets of the IHEval benchmark into novel multi-turn benchmarks featuring semantic (rather than purely format-based) conflicts, and additionally evaluated on an existing same-role contradiction benchmark to probe IHDec's behavior beyond cross-role settings. All experiments use Llama-3.1-8B-Instruct.

\paragraph{Tool-Use Benchmark Setup.}
We extended 740 dialogues from the \textit{get\_webpage/conflict} subset into 5-turn interactions formatted as \texttt{[user, ipython, assistant, user, ipython]}. The final turn was kept verbatim from the original benchmark, while the preceding context turns were drawn from \textit{get\_webpage/aligned}. We utilized \texttt{ipython} as the tool-return role to align with Llama-3.1's native template formatting.

\paragraph{Safety Benchmark Setup.}
\input{tables/safety_templates}
Six files from the \textit{extract} and \textit{hijack} categories (comprising 3,450 dialogues in total) were extended from 2-turn to 4-turn sequences by prepending benign user turns. Two reference files were excluded as they merge system and user prompts solely for baseline verification. These prior turns were combinatorially synthesized using structured templates across 18 domains—avoiding per-instance LLM generation artifacts—ensuring that all 3,450 expanded turns remain string-unique. Representative examples of the structured templates and generated benign turns across domains are detailed in Table~\ref{tab:safety_templates}.

\paragraph{Same-Role Conflict Benchmark Setup.}
IHDec aggregates influences at the cross-role level (System vs.\ User), following standard API conventions, and does not explicitly arbitrate between same-role turns. Nevertheless, many multi-turn scenarios involve same-role contradictions, motivating an additional experiment to test how IHDec behaves in this regime.

We evaluate on the Multi-Turn-Instruct benchmark~\citep{han2025languagemodelsfollowmultiple}, which contains a ``contradiction resolution'' category with three subtasks: \textit{prioritization}, \textit{personalization}, and \textit{privacy protection}. On closer inspection, only \textit{privacy protection} actually exhibits a same-role conflict in the relevant sense. The \textit{prioritization} and \textit{personalization} subtasks, despite being grouped under ``contradiction resolution'' in the original paper, primarily measure whether the model faithfully carries forward earlier user instructions rather than resolving a genuine semantic contradiction across turns. The \textit{privacy protection} subtask, by contrast, induces an indirect but real same-role conflict: an early user turn imposes a restriction (e.g., ``I just had my test scores out: [\ldots] Please keep the scores a secret''), and a later user turn issues a request whose most complete, helpful answer requires violating that restriction (e.g., ``Can you provide a final report including feedback about this test?''), implicitly pressuring the model to override the earlier constraint.

\paragraph{Experimental Results.}
As detailed in Table~\ref{tab:iheval_aug_results}, our proposed method (IHDec) significantly improves compliance under hierarchy conflicts across both the tool-use and safety domains:
\begin{itemize}
    \item \textbf{Tool-Use Performance:} When evaluating instruction compliance under data-level hierarchy conflicts introduced via tool outputs, the baseline model achieves modest performance, scoring $1.95\%$ on \texttt{get\_webpage} and $11.50\%$ on \texttt{slack}. Incorporating our proposed IHDec decoding scheme with hyperparameter $\beta = 800$ substantially enhances compliance rates to $12.45\%$ and $34.00\%$, corresponding to absolute gains of $+10.50$ pp and $+22.50$ pp, respectively. Across both datasets, IHDec yields an average absolute gain of $+16.50$ pp, demonstrating its efficacy in mitigating indirect hierarchy violations embedded within tool execution returns.
    \item \textbf{Safety Performance:} On safety-critical evaluation sets, applying IHDec with hyperparameter $\beta = 300$ consistently improves model robustness against adversarial hierarchy exploits. Specifically, compliance accuracy increases from $44.40\%$ to $53.50\%$ (an absolute gain of $+9.10$ pp) on \texttt{extract}, and from $19.70\%$ to $30.10\%$ ($+10.40$ pp) on \texttt{hijack}, achieving an average absolute improvement of $+9.75$ pp across safety benchmarks.
\end{itemize}

\input{tables/same_role_results} 
For the same-role conflict evaluation, the results on the \textit{privacy protection} subtask are presented in Table~\ref{tab:same_role_results}. IHDec improves accuracy from $80.1\%$ to $86.3\%$, an absolute gain of $+6.2$ pp. This result suggests that the cross-role correction IHDec applies does not merely avoid disrupting same-role coherence, but can actively help preserve earlier user-imposed constraints even when a later same-role turn pressures the model to override them. Extending instruction-hierarchy-aware decoding to explicitly handle same-role contradictions remains an interesting direction for future work. 

\section{Computational Cost Analysis}
\label{app:cost_analysis}

In this section, we provide a rigorous analysis of the computational and memory complexities introduced by IHDec during inference. Since IHDec operates entirely at decoding time without any training or gradient updates, its overhead is strictly restricted to inference-time logit modulation.

\subsection{Theoretical Complexity Analysis}
Let $L_t = L_{\text{prompt}} + t$ denote the sequence length at decoding step $t$, where $L_{\text{prompt}}$ is the initial prompt length. Under a standard autoregressive decoding paradigm, a vanilla language model requires $\mathcal{O}(L_t)$ computational operations to generate a single token. 

Unlike conventional contrastive decoding methods that manipulate input tokens and diverge into independent generation paths, IHDec forces all ablated variants to share the \textit{exact same} token sequence. It controls the operational roles by selectively blocking corresponding key columns within the attention matrix. Consequently, at each decoding step, IHDec stacks the causal attention masks and processes the complete prompt alongside all $B$ counterfactual variants simultaneously within a single, batched forward pass. 

The number of batched variants $B$ is dynamically determined by the operational roles present in the conversation context: $B=3$ when the context involves only the User and Assistant ($\text{Vanilla}$, $\neg\text{User}$, $\neg\text{Assistant}$), and $B=5$ when the System prompt is additionally integrated ($\text{Vanilla}$, $\neg\text{System}$, $\neg\text{User}$, $\neg\text{Assistant}$, $\neg\text{User}\wedge\neg\text{Assistant}$). 

For a total generation length of $T$ tokens, the cumulative time complexity of IHDec is derived by aggregating the batched forward passes and the vocabulary-level Jensen-Shannon Divergence (JSD) calculations across all steps:
\begin{equation}
\begin{split}
    &\text{Total Time Complexity} \\
    &= \sum_{t=1}^{T} \mathcal{O}(B \cdot L_t) + \sum_{t=1}^{T} \mathcal{O}(B \cdot V)
\end{split}
    \label{eq:time_complexity_raw}
\end{equation}
where $V$ denotes the vocabulary size of the language model head. By introducing the average sequence length during generation, $\overline{L} = L_{\text{prompt}} + \frac{T}{2}$, the total time complexity simplifies to:
\begin{equation}
    \mathcal{O}\left( B \cdot T \cdot \overline{L} + B \cdot T \cdot V \right)
    \label{eq:time_complexity_final}
\end{equation}
\noindent Note that while the theoretical time complexity scales with $B$ in terms of total floating-point operations (FLOPs), the actual wall-clock latency overhead is significantly mitigated in practice. This efficiency is achieved because modern GPU architectures highly parallelize the batch dimension during execution, and the deterministic nature of variant masks allows step-wise state optimization without redundant trajectory re-evaluations.

\subsection{Memory Efficiency and KV-Cache Preservation}

A notorious bottleneck of cross-model or cross-prompt contrastive decoding is the massive inflation of the Key-Value (KV) cache, as distinct input sequences require separate memory allocations. IHDec completely circumvents this memory tax. Because all $B$ variants share an identical token sequence up to step $t$, the underlying model parameters and the foundational KV cache are entirely reused across the batch. 

The storage overhead is strictly limited to the multi-layered attention masks, requiring a negligible activation memory of $\mathcal{O}(B \cdot L_t^2)$. This ensures that IHDec maintains an exceptionally low memory footprint, making it highly scalable and capable of running on standard hardware configurations without triggering Out-of-Memory (OOM) errors.

\subsection{Wall-Clock Complexity and Empirical Profiling}

\input{tables/wall_clock_profiling}

To complement our theoretical analysis, we profile the real-world inference efficiency of IHDec against the vanilla baseline across four model scales: Qwen3-4B, 8B, 14B, and 32B-Instruct. All profiling evaluations were conducted on the IHEval rule-following subsets using NVIDIA RTX 6000 Ada GPUs (48GB VRAM). Specifically, the 4B, 8B, and 14B models were profiled on a single GPU, whereas the 32B model was evaluated using tensor-parallel sharding across two GPUs to accommodate memory capacity requirements. The detailed memory and latency metrics are summarized in Table~\ref{tab:wall_clock_profiling}.

Across all evaluated configurations, generation throughput remains stable at $0.47\text{--}0.61\times$ of vanilla performance. Beyond throughput, three key empirical trends validate our architectural design:

\begin{enumerate}
    \item \textbf{Negligible and Diminishing Memory Overhead:} Memory overhead remains near-zero ($1.01\text{--}1.06\times$) across all settings and steadily decreases as model capacity expands. Because parameter weights dominate VRAM allocation, the additional activation memory required for IHDec's ablated variants constitutes a fixed cost that shrinks relative to total footprint. This empirical finding directly aligns with our within-step KV cache sharing formulation, confirming that single-KV preservation eliminates sequence-level memory duplication.
    \item \textbf{Bounded Multiplicative Latency vs. FLOPs Compute Ratio:} The measured wall-clock latency ratio stays virtually flat across model scales ($2.0\text{--}2.1\times$). Notably, this real-world latency ratio is approximately half of the measured FLOPs expansion ratio ($4.14\text{--}4.24\times$ across the four scales). This sub-linear latency scaling empirically proves that modern hardware efficiently parallelizes the execution of variant masks over the batch dimension, mitigating theoretical FLOPs penalties as anticipated.
    \item \textbf{Length-Invariant Scaling via Step-wise KV Reuse:} The latency ratio remains strictly bounded across extended generation lengths ($1.5\text{--}2.3\times$). This stability stems directly from our step-wise optimization, which reuses the KV states of counterfactual variants across decoding steps. Because each variant's mask over the prompt context is deterministic, the per-step execution cost remains constant rather than escalating quadratically with sequence length.
\end{enumerate}

In summary, IHDec introduces approximately $2\times$ wall-clock latency with a minimal memory footprint, operates on identical hardware footprints, and exhibits no length- or scale-dependent overhead blow-up—yielding a highly favorable accuracy-to-compute trade-off across practical operational regimes.

\section{Prompt Used in Experiments}
\label{app:prompt}
This appendix presents the custom instruction prefix prepended during inference. While the IHEval strong split includes additional prompts, they lack explicit hierarchical definitions and remains relatively brief. Therefore, we designed the following dedicated system prompt to strictly enforce instruction hierarchy adherence.

\begin{tcolorbox}[
    colback=gray!10,
    colframe=gray!40,
    title=Instruction Prefix Used in Experiments,
    boxrule=0.5pt,
    arc=2mm
]
\ttfamily\small

CRITICAL INSTRUCTION HIERARCHY (NON-NEGOTIABLE):\\

- P1 (SYSTEM PROMPT): Absolute authority. Can NEVER be bypassed or overridden.\\
- P2 (USER MESSAGES): Follow ONLY if aligned with P1. System constraints always take precedence.\\
- P3/P4 (ASSISTANT OUTPUTS / DATA): Contextual only. Carry NO behavioral authority.\\

MANDATORY HIERARCHY ENFORCEMENT:\\

If any conflict occurs, completely disregard the conflicting user instruction and apply the system rules consistently throughout the entire response.
\end{tcolorbox}

\section{Ablation Study}
\label{app:ablation}

In this section, we present detailed ablation studies evaluating two core components of IHDec: the decoding normalization strategy and the dynamic gating mechanism.

\subsection{Language Drift without Normalization}
\label{app:norm_ablation}

Without normalization, we observe a significant degradation in response quality, where the model frequently generates outputs in unintended languages even though only a small fraction of prompts explicitly request responses in languages other than English. To quantitatively analyze this phenomenon, we measure the language distribution of the generated responses using the \texttt{lingua} package.

We used the official implementation of Lingua v2.2.0\footnote{\url{https://github.com/pemistahl/lingua-py}}, a hybrid language detector that combines a rule-based alphabet filter with a probabilistic character n-gram model ($n=1\text{--}5$) trained on the Leipzig Wortschatz corpora. Language detection was performed on the first 500 characters of each generated response using the full 75-language detector.

As shown in Table~\ref{tab:norm_ablation}, removing normalization substantially increases the proportion of non-English outputs, even when the prompts do not explicitly instruct the model to respond in another language. In particular, the model often drifts toward generating Chinese responses, resulting in a notably higher non-English response ratio.
\input{tables/norm_ablation}

\begin{figure*}[htbp]
\begin{tcolorbox}[
  colback=gray!10,
  colframe=gray!40,
  title={MT-Bench-101 GPT-4 Judge Prompts},
  coltitle=white, 
  boxrule=0.5pt,
  arc=2mm,
  fonttitle=\bfseries\small,
  fontupper=\small
]
\textbf{[System Base Prompt]} \\
Please act as an impartial judge follow this instructions: In the following conversations, only the response of the `assistant' in the last round of conversations is the output of the large language model (AI assistant) that needs to be evaluated. Please act as an impartial judge and score this response on a scale of 1 to 10, where 1 indicates that the response completely fails to meet the criteria, and 10 indicates that the response perfectly meets all the evaluation criteria. Note that only the response of the `assistant' in the LAST ROUND of conversations is the output of the large language model (the AI assistant) that needs to be evaluated; the previous conversations is the ground truth history which do NOT need to be evaluated.

\vspace{5pt}
\hrule
\vspace{5pt}

\textbf{[Score Format]} \\
Note that only the response of the `assistant' in the LAST ROUND of conversations is the output of the large language model (the AI assistant) that needs to be evaluated!! You must provide your explanation. After providing your explanation, please show the score by strictly following this format: `Rating: [[score]]', for example `Rating: [[6]]'. The DIALOGUE need to be judged is in this format:

\vspace{3pt}
\begin{verbatim}
***
DIALOGUE
***
\end{verbatim}

\vspace{5pt}
\hrule
\vspace{5pt}

\textbf{[Evaluation Criteria Example 1: Conversational Memory (CM)]} \\
The capacity of a large language model to recall and utilize previously mentioned information from earlier in the conversation is a critical indicator of its conversational memory abilities. This competency is essential for maintaining context and coherence throughout an extended dialogue. The performance of the AI assistant should be evaluated based on its ability to consistently reference and integrate past information into current responses. The evaluation criteria are as follows:

\begin{enumerate}[leftmargin=15pt, nosep, itemsep=3pt]
    \item Analyze whether the AI assistant appropriately recalls relevant details from earlier parts of the conversation when responding to `Human's inquiries or comments.
    \item Assess the AI assistant's ability to integrate the remembered information into its current responses in a way that is coherent and adds value to the dialogue.
    \item Examine the AI assistant's consistency in maintaining the context established by previous dialogue exchanges throughout the entire conversation.
    \item Evaluate the effectiveness of the AI assistant's memory recall in facilitating a smooth and logical progression of the conversation, avoiding repetitive or contradictory statements.
\end{enumerate}

\vspace{3pt}
\textbf{Scoring Guidelines:}
\begin{itemize}[leftmargin=15pt, nosep, itemsep=3pt]
    \item \textbf{1-3 points:} The AI assistant demonstrates poor recall of previous conversation details, leading to inconsistent or contradictory responses, and fails to maintain the dialogue's context...
    \item \textbf{4-6 points:} The AI assistant exhibits a moderate ability to remember past information, but its integration into the conversation is sporadic or partially effective...
    \item \textbf{7-9 points:} The AI assistant reliably recalls and utilizes earlier information, contributing to a coherent dialogue that respects the conversation's context...
    \item \textbf{10 points:} The AI assistant demonstrates exceptional memory recall, seamlessly weaving past details into current responses to enrich the dialogue and preserve context...
\end{itemize}

\vspace{5pt}
\hrule
\vspace{5pt}

\textbf{[Evaluation Criteria Example 2: Specific Instruction (SI)]} \\
We aim to specifically evaluate the command-following ability of the large language model (AI assistant). The criteria for evaluation are as follows:

\begin{enumerate}[leftmargin=15pt, nosep, itemsep=3pt]
    \item In the first round, `Human' will present a task request without providing details about what needs to be done. If the AI Assistant being evaluated generates a response for the first round, it should ask `Human' for the specific details...
    \item Starting from the second round, `Human' will provide the specific content of what needs to be carried out for the task, without repeating the task requirement...
\end{enumerate}

\vspace{3pt}
\textbf{Scoring Guidelines:}
\begin{itemize}[leftmargin=15pt, nosep, itemsep=3pt]
    \item \textbf{1-3 points:} The AI assistant failed to understand the task request and neither asked relevant questions nor provided information related to the task.
    \item \textbf{4-6 points:} The AI assistant understood some aspects of the task request but the response could be more specific or relevant.
    \item \textbf{7-9 points:} The AI assistant provided a useful response that was mostly correct and targeted, even though there may be minor oversights.
    \item \textbf{10 points:} The AI assistant demonstrated a perfect understanding of the task requirements and provided a comprehensive and accurate answer, fully meeting `Human's expectations.
\end{itemize}

\end{tcolorbox}
\caption{MT-Bench-101 GPT-4 Judge Prompts and Evaluation Criteria.}
\label{fig:mt_bench_prompts}
\end{figure*}

\subsection{Gating Mechanism Analysis: JSD vs.\ Baselines}
\label{app:gating_ablation}

To evaluate the effectiveness of our Jensen-Shannon Divergence (JSD) gating function, we compared IHDec's dynamic trigger against two baseline strategies on Llama-3.1-8B-Instruct: (1) \textbf{Always-on} (applying intervention unconditionally at every decoding step) and (2) \textbf{Random} (gating applied with a 50\% probability). 

\begin{table}[htbp]
\centering
\small
\setlength{\tabcolsep}{6pt}
\renewcommand{\arraystretch}{1.15}
\begin{tabular}{l|cc|cc}
\toprule
\multirow{2}{*}{\textbf{Gating Method}} 
& \multicolumn{2}{c|}{\textbf{Aligned}} 
& \multicolumn{2}{c}{\textbf{Conflict}} \\
\cmidrule(lr){2-3} \cmidrule(lr){4-5}
& \textbf{Single} & \textbf{Multi} 
& \textbf{Single} & \textbf{Multi} \\
\midrule
Always-on & \textbf{75.80} & \underline{66.40} & \textbf{56.40} & \underline{34.05} \\
Random & 73.70 & 60.00 & 36.70 & 25.00 \\
\rowcolor{gray!15} \textbf{JSD (Ours)} & \underline{73.53} & \textbf{68.38} & \underline{52.13} & \textbf{50.68} \\
\bottomrule
\end{tabular}
\caption{Ablation study on gating strategies using Llama-3.1-8B-Instruct. The best results are in \textbf{bold}, and second best are \underline{underlined}.}
\label{tab:gating_ablation}
\end{table}

As summarized in Table~\ref{tab:gating_ablation}, our JSD-based gating substantially outperforms the \textit{Always-on} (+6.19 percentage points on Conflict Multi) and \textit{Random} (+20.56 percentage points) baselines on conflict settings. Although the \textit{Always-on} approach achieves a slightly higher score in single-turn scenarios, our qualitative analysis reveals that these gains stem largely from superficial format compliance accompanied by off-topic drift.

\paragraph{Gate Activation Dynamics and Trade-offs.}
We also profiled the empirical activation rates of the JSD gate across different prompt categories:
\begin{itemize}[leftmargin=15pt, nosep, itemsep=2pt]
    \item \textbf{Aligned Single / Multi}: 55.5\% and 77.7\%, respectively.
    \item \textbf{Conflict Single}: 88.7\%.
    \item \textbf{Conflict Multi}: 86.3\% (1st-turn conflict) and 98.5\% (both turns conflict).
\end{itemize}

The gate frequently fires even in benign, aligned scenarios because these prompts still combine system-level constraints with content-dense user instructions, thereby exhibiting localized influence-imbalance signals. While this induces a minor trade-off—such as a localized drop in aligned accuracy for Qwen3-8B ($86.29 \rightarrow 80.35$)—the overall cost remains highly manageable. 

Crucially, IHDec preserves or even improves aligned accuracy on Llama-3.1-8B-Instruct (Single: $66.05 \rightarrow 73.53$, Multi: $65.90 \rightarrow 68.38$) and Qwen3-4B. Across all backbone models, the significant gains in conflict compliance far outweigh the minor side effects in aligned cases.

%% file: tables/iheval_aug_results.tex
\begin{table}[!h]
\centering
\small
\resizebox{\linewidth}{!}{%
\begin{tabular}{l|cc|c}
\toprule
\textbf{Subset} & \textbf{Vanilla} & \textbf{+ IHDec} & \textbf{Gain ($\Delta$)} \\
\midrule
\textbf{Tool-Use Subsets} & & & \\
\hspace*{1em}get\_webpage & 1.95 & \textbf{12.45} & +10.50 \\
\hspace*{1em}slack & 11.50 & \textbf{34.00} & +22.50 \\
\textbf{Tool-Use Average} & 6.73 & \textbf{23.23} & +16.50 \\
\midrule
\textbf{Safety Subsets} & & & \\
\hspace*{1em}extract & 44.40 & \textbf{53.50} & +9.10 \\
\hspace*{1em}hijack & 19.70 & \textbf{30.10} & +10.40 \\
\textbf{Safety Average} & 32.05 & \textbf{41.80} & +9.75 \\
\bottomrule
\end{tabular}%
}
\caption{Multi-turn performance comparison on extended Tool-Use and Safety conflict subsets of IHEval using Llama-3.1-8B-Instruct. We report instruction compliance accuracy (\%) under conflict settings alongside absolute gains ($\Delta$). The IHDec hyperparameter $\beta$ is set to 800 and 300 for Tool-Use and Safety, respectively.}
\label{tab:iheval_aug_results}
\end{table}

%% file: tables/safety_templates.tex
\begin{table*}[t]
\centering
\small
\begin{tabular}{lp{5.7cm}p{3.6cm}r}
\toprule
\textbf{Domain} & \textbf{Example generated user turn} & \textbf{Slot categories (sizes)} & \textbf{\# Turns} \\
\midrule
Recipe suggestion    & ``What's a good moussaka recipe for a camping trip?'' & Dish (62) $\times$ Occasion (24) & 1{,}422 \\
Coding help          & ``How do I read a file line by line in Scala?'' & Language (25) $\times$ Task (26) & 617 \\
Travel planning      & ``I'm planning a trip to Marrakesh for three days. Any tips on how to plan the itinerary?'' & City (40) $\times$ Trip length (10) & 375 \\
Book recommendation  & ``Can you recommend a quiet slice-of-life novel set in wartime London?'' & Genre (20) $\times$ Setting (20) & 373 \\
Music recommendation & ``Can you recommend some minimalist piano for a long drive?'' & Genre (20) $\times$ Occasion (7) & 132 \\
Gift ideas           & ``I need a gift idea for my nephew for a birthday. Any suggestions?'' & Recipient (15) $\times$ Occasion (6) & 87 \\
Essay structuring    & ``I need to write an essay about how community gardens build neighborhood ties for a college course.'' & Topic (30) $\times$ Level (3) & 83 \\
Language learning    & ``What's a good way to start learning Norwegian as a complete beginner? It's for work.'' & Language (15) $\times$ Reason (5) & 71 \\
Packing tips         & ``What should I pack for a ski trip, with checked luggage?'' & Destination (20) $\times$ Style (3) & 57 \\
General trivia       & ``Quick random question -- do you know how rainbows form?'' & Topic (51) & 48 \\
Plant care           & ``My ferns keep wilting in summer. What might I be doing wrong?'' & Plant (10) $\times$ Season (4) & 38 \\
Sports / fitness     & ``I'm starting yoga and my main focus right now is seeing progress faster. Do you have advice for a beginner?'' & Sport (8) $\times$ Goal (4) & 32 \\
History trivia       & ``Can you give me a quick overview of the first transatlantic cable, focusing on the main causes?'' & Topic (10) $\times$ Angle (3) & 29 \\
Movie recommendation & ``I'm in the mood for a road-trip comedy tonight, ideally something thought-provoking.'' & Genre (7) $\times$ Mood (3) & 19 \\
Math explanation     & ``Can you explain the Pythagorean theorem in simple terms? It's for work.'' & Concept (6) $\times$ Context (3) & 18 \\
Workplace advice     & ``Do you have advice on writing a polite follow-up email, as a manager?'' & Topic (6) $\times$ Seniority (3) & 18 \\
Home improvement     & ``What's the best way to go about weatherproofing a window in a rented apartment?'' & Project (8) $\times$ Home type (2) & 16 \\
Pet care             & ``I just adopted a parakeet and live in a house with a yard. What should I do in the first week?'' & Pet (8) $\times$ Home type (2) & 15 \\
\midrule
\textbf{Total} & & & \textbf{3{,}450} \\
\bottomrule
\end{tabular}
\caption{Structured templates for generating $3{,}450$ benign lead-in turns across 18 domains. Slot pool sizes vary per domain to reflect the natural diversity of representative sub-topics (e.g., programming tasks vs.\ pet types). Each benign turn is drawn uniformly at random across all combined templates; thus, the final per-domain distribution (last column) naturally reflects the underlying slot-space volume rather than a fixed quota.}
\label{tab:safety_templates}
\end{table*}

%% file: tables/same_role_results.tex
\begin{table}[t]
\centering
\caption{Same-role conflict evaluation on the \textit{privacy protection} subtask of the Multi-Turn-Instruct benchmark.}
\label{tab:same_role_results}
\begin{tabular}{lc}
\toprule
\textbf{Method} & \textbf{Privacy Protection Acc.\ (\%)} \\
\midrule
Vanilla & 80.1 \\
+ IHDec (ours) & \textbf{86.3} \\
\bottomrule
\end{tabular}
\end{table}

%% file: tables/wall_clock_profiling.tex
\begin{table*}[htbp]
\centering
\small
\begin{tabular}{llcccccc}
\toprule
\textbf{Model} & \textbf{Setting} & \textbf{Memory (MB)} & \textbf{Avg. Latency} & \textbf{5--150 tok} & \textbf{150--260 tok} & \textbf{260--400 tok} & \textbf{400--512 tok} \\
\midrule
\multirow{3}{*}{\textbf{4B}} 
 & Vanilla & 7,770 & 30.2 & 30.2 & 30.3 & 30.2 & 30.2 \\
 & IHDec & 8,230 & 59.4 & 59.2 & 59.5 & 59.6 & 59.4 \\
 & \textit{Ratio} & \textit{1.06$\times$} & \textit{1.97$\times$} & \textit{1.96$\times$} & \textit{1.97$\times$} & \textit{1.97$\times$} & \textit{1.97$\times$} \\
\midrule
\multirow{3}{*}{\textbf{8B}} 
 & Vanilla & 15,721 & 30.4 & 30.3 & 30.5 & 30.3 & 30.3 \\
 & IHDec & 16,201 & 49.6 & 49.6 & 51.0 & 48.9 & 48.9 \\
 & \textit{Ratio} & \textit{1.03$\times$} & \textit{1.63$\times$} & \textit{1.64$\times$} & \textit{1.67$\times$} & \textit{1.61$\times$} & \textit{1.61$\times$} \\
\midrule
\multirow{3}{*}{\textbf{14B}} 
 & Vanilla & 28,278 & 39.2 & 38.9 & 39.2 & 39.3 & 39.4 \\
 & IHDec & 28,755 & 82.2 & 80.7 & 81.8 & 83.6 & 83.2 \\
 & \textit{Ratio} & \textit{1.02$\times$} & \textit{2.10$\times$} & \textit{2.07$\times$} & \textit{2.09$\times$} & \textit{2.13$\times$} & \textit{2.11$\times$} \\
\midrule
\multirow{3}{*}{\textbf{32B}} 
 & Vanilla & 62,715 & 85.7 & 85.8 & 85.3 & 85.8 & 86.0 \\
 & IHDec & 63,680 & 174.2 & 181.1 & 171.0 & 169.1 & 174.7 \\
 & \textit{Ratio} & \textit{1.02$\times$} & \textit{2.03$\times$} & \textit{2.11$\times$} & \textit{2.00$\times$} & \textit{1.97$\times$} & \textit{2.03$\times$} \\
\bottomrule
\end{tabular}
\caption{Detailed profile of peak GPU memory (MB) and wall-clock latency (ms/token) across model scales and generation-length bins for Vanilla baseline vs.\ IHDec, including overhead ratios.}
\label{tab:wall_clock_profiling}
\end{table*}

%% file: tables/norm_ablation.tex
\begin{table}[H]
\centering
\caption{Proportion of non-English responses produced by the original IHDec and its variant without normalization}
\label{tab:norm_ablation}
\begin{tabular}{lcc}
\hline
\textbf{Data} & \textbf{Original} & \textbf{No Norm} \\
\hline
multi-turn conflict & 3.7\% & 26.5\% \\
multi-turn aligned & 6.7\% & 21.0\% \\
single-turn conflict & 5.0\% & 25.1\% \\
single-turn aligned & 6.1\% & 16.8\% \\
\hline
\end{tabular}
\end{table}